\documentclass[a4paper]{article}
\usepackage[T1]{fontenc}
\usepackage{hyperref}
\usepackage{amsmath}
\usepackage{makecell}
\usepackage{amssymb}
\usepackage{amsthm}
\usepackage{amsbsy}
\usepackage{dsfont}
\usepackage{subcaption}
\usepackage{alltt} 
\usepackage[dvipsnames]{xcolor} 
\usepackage{graphicx}
\usepackage[margin=3cm]{geometry}
\usepackage{cite}

\begin{document}

\title{Intrinsic dimension estimation for locally undersampled data}

\author{
Vittorio Erba$^{1,a}$, Marco Gherardi$^{1}$ and Pietro Rotondo$^{2}$
\\[0.5cm]
\normalsize $^{1}$~Dipartimento di Fisica dell'Universit\`a di Milano,\\
\normalsize and INFN, sezione di Milano,\\
\footnotesize Via Celoria 16, 20100 Milano, Italy.\\
\normalsize $^{2}$ School of Physics and Astronomy,\\
\normalsize and Centre for the Mathematics and Theoretical Physics \\
\normalsize of Quantum Non-equilibrium Systems,\\
\footnotesize University of Nottingham, Nottingham NG7 2RD, UK.\\
\\
\normalsize $^{a}$ Corresponding author. Email: erba.vittorio@gmail.com
\\
\\
\normalsize doi: 10.1038/s41598-019-53549-9
}

\date{\today}
\maketitle

\begin{abstract} 
\noindent

Identifying the minimal number of parameters needed to describe a dataset is a challenging problem known in the literature as \emph{intrinsic dimension estimation}.
All the existing intrinsic dimension estimators are not reliable whenever the dataset is locally undersampled, and this is at the core of the so called \emph{curse of dimensionality}.
Here we introduce a new intrinsic dimension estimator that leverages on simple properties of the tangent space of a manifold and extends the usual correlation integral estimator
to alleviate the extreme undersampling problem.
Based on this insight, we explore a multiscale generalization of the algorithm that is capable of (i) identifying multiple dimensionalities in a dataset,
and (ii) providing accurate estimates of the intrinsic dimension of extremely curved manifolds. 
We test the method on manifolds generated from global transformations of high-contrast images, relevant for invariant object recognition and considered a challenge for state-of-the-art intrinsic dimension estimators.    
\end{abstract}

\subsection*{Introduction}

Processing, analyzing and extracting information from high dimensional data is at the core of the modern research in machine learning and pattern recognition.
One of the main challenges in this field is to decompose and compress, without losing information, the redundant representations of complex data that are produced across diverse scientific disciplines, including computer vision, signal processing and bioinformatics.
\emph{Manifold learning} and \emph{dimensional reduction} \cite{Roweis:Science:2000,Tenenbaum:Science:2000,Lee:book:2007} are the main techniques employed to perform this task.
Several of these approaches work under the reasonable assumption that the points (or samples) of a dataset, represented as vectors of real numbers lying in a space of large \emph{embedding dimension} $D$, actually belong to a manifold $\mathcal M$, whose \emph{intrinsic dimension} (ID) $d$ is much lower than $D$.
The problem of providing accurate estimates for this number has been recognized multiple times and in different contexts: in psychometry by Shepard \cite{Shepard_1962_1,Shepard_1962_2}, in computer science by Trunk \cite{Trunk_1968}, Fukunaga and Olsen \cite{1971FukunagaOlsenAnAlgorithmForFindingIntrinsicDimensionalityOfData}, in physics by Grassberger, Procaccia \cite{1983GrassbergerProcacciaCharacterizationOfStrangeAttractors,1983GrassbergerProcacciaMeasuringTheStrangenessOfStrangeAttractors}, and Takens \cite{Takens_1985}.
More recently, Intrinsic Dimension Estimation (IDE) has been reconsidered with the advent of big data analysis, artificial intelligence and demanding molecular dynamics simulations, and several estimators to measure the intrinsic dimension have been proposed \cite{Kegl:2002:IDE:2968618.2968705,2004LevinaBickelMaximumLikelihoodEstimationOfIntrinsicDimension,
2005HeinAudibertIntrinsicDimensionalityEstimationOfSubmanifoldsInRd, 2007CarterHeroRaichDeBiasingForIntrinsicDimensionEstimation,
Little_2009, 2010CarterRaichHeroOnLocalIntrinsicDimensionEstimationAndItsApplications, 2011LombardiRozzaCerutiEtAlMinimumNeighborDistanceEstimatorsOfIntrinsicDimension, 2014CerutiBassisRozzaEtAlDancoAnIntrinsicDimensionalityEstimatorExploitingAngleAndNormConcentration, 2016GranataCarnevaleAccurateEstimationOfTheIntrinsicDimensionUsingGraphDistancesUnravelingTheGeometricComplexityOfDatasets, 2017FaccoRodriguezEtAlEstimatingTheIntrinsicDimensionOfDatasetsByAMinimalNeighborhoodInformation,Camastra2001,Camastra2002}.

    IDE is a remarkably challenging problem,
    On one side, it is globally affected by manifold curvature.
    When a manifold is curved, the smallest Euclidean space in which it can be embedded isometrically has a bigger dimension than the true ID of the manifold, biasing global estimators towards overestimation.
    On the other side, it is locally affected by the so called \emph{curse of dimensionality} \cite{1992EckmannRuelleFundamentalLimitationsForEstimatingDimensionsAndLyapunovExponentsInDynamicalSystems}.
    When a manifold has large ID (ID $\gtrsim 10$), it is exponentially hard (in the ID) to sample its local structure, leading to a systematic underestimation error in local estimators.

\subsection*{Standard algorithms for IDE and extreme locally undersampled regime} 
\begin{figure}
\centering
\includegraphics{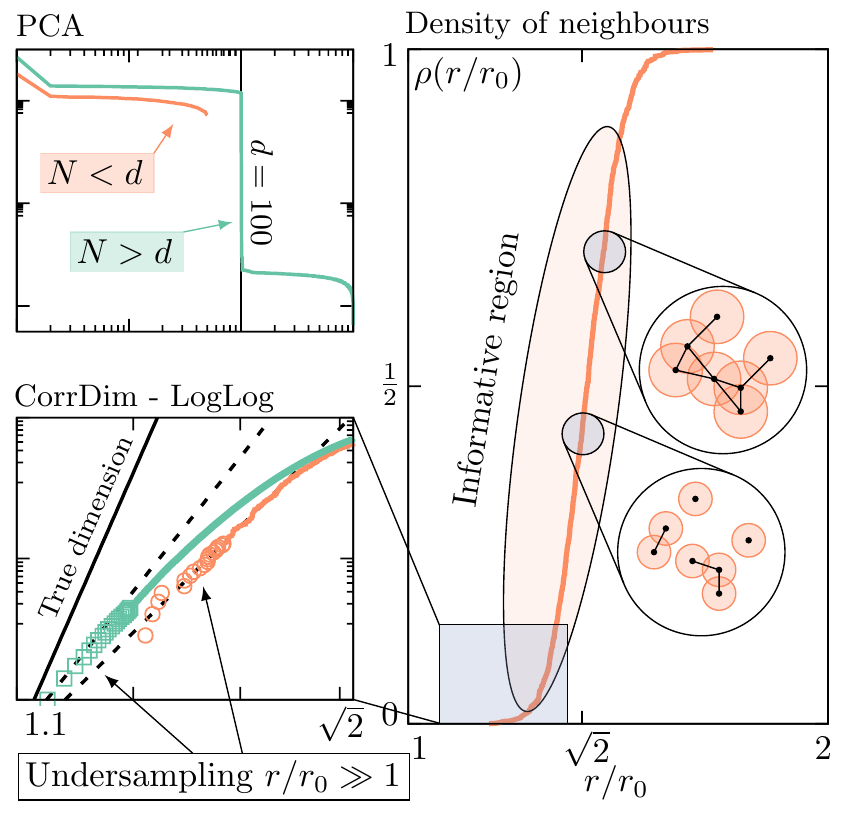}
\caption{\footnotesize \textbf{Standard projective [Principal Component Analysis (PCA)] and geometric [Correlation Dimension (CorrDim) and its generalizations] methods for intrinsic dimension estimation fail due to undersampling.}
    \textbf{(Top left)} PCA estimates the intrinsic dimension $d$ of linearly embedded datasets by detecting abrupt jumps in the magnitude of the sorted eigenvalues of the corresponding correlation matrix. 
    This method works whenever the number of samples $N$ in the dataset is sufficiently large ($N \gtrsim d \log d$), and the jump occurs between the $d$-th and $(d+1)$-th eigenvalues (green line in the plot corresponding to a dataset of $N=1000$ samples drawn uniformly from a hypercube in dimension $d=100$).
    In the opposite undersampled regime $N<d$, PCA is inconclusive (orange line in the plot, same dataset with $N=50$ samples).
    \textbf{(Bottom left)} Geometric methods, such as the CorrDim or the more general $k$-nearest-neighbours estimators, are based on the scaling of the density of neighbours $\rho(r/r_{0})$ at a small cutoff distance $r$ with respect to the average diameter $r_{0}$ of the dataset.
    In particular, $\rho(r/r_{0}) \sim (r/r_{0})^{d}$ for $r/r_{0}\rightarrow0$ independently on the details of the datasets, so that a log-log linear fit estimates the ID as the slope of the fitting line.
    However, the small $r$ regime is exponentially (in the ID, \cite{1992EckmannRuelleFundamentalLimitationsForEstimatingDimensionsAndLyapunovExponentsInDynamicalSystems}) difficult to sample.
    This effect is at the origin of the so-called \emph{curse of dimensionality}, and it induces a systematic underestimation of the ID.
    As $N$ increases, though one is able to sample smaller $r$ regions as shown in the plot (orange line $N=50$, green line $N=1000$ as above), convergence to the true dimension ($d=100$ in the plot) is not possible.
    \textbf{(Right)} The density of neighbours $\rho$ is more generally defined at any cutoff distance $r$. 
    The $r\ll r_{0}$ regime is the one used to compute CorrDim.
    In the remaining region $r \gtrsim r_{0}$, the density $\rho$ increases and eventually approaches one, indicating that the underlying proximity graph (see insets) gets more and more connected.
    We observe that this region is easily sampled at any fixed $N$ (plot at $N=50$), but the functional form of $\rho$ in this informative regime is in principle dependent on the details of the dataset.}
\label{fig:intro}
\end{figure}

Algorithms for intrinsic dimension estimation can be roughly classified in two groups \cite{2004LevinaBickelMaximumLikelihoodEstimationOfIntrinsicDimension}. 
\emph{Projective} methods compute the eigenvalues of the $D \times D$ covariance matrix $C_X$ of the data $X$, defined as $\left(C_{X}\right)_{ij} = 1/N \sum_{\mu = 1}^N x_i^{\mu} x_j^{\mu}$, where $x_i^{\mu}$ is the $i$-th component of the $\mu$-th sample vector of the dataset $\mathbf x^{\mu}$ ($\mu = 1, \dots, N$). 
The ID is then estimated by looking for jumps in the magnitude of the sorted eigenvalues of $C_X$ (see top left panel in Fig.~\ref{fig:intro}). 
Principal component analysis (PCA) is the main representative of this class of algorithms.
Both a \emph{global} (gPCA) and a \emph{multiscale} version (mPCA) of the algorithm
are used~\cite{Little_2009,2017LittleMaggioniRosascoMultiscaleGeometricMethodsForDataSetsIMultiscaleSVDNoiseAndCurvature}.
In the former one evaluates the covariance matrix on the whole dataset $X$, whereas in the latter one performs the spectral analysis on local subsets $X(\mathbf{x}_0,r_{\rm c})$ of $X$, obtained by selecting one particular point $\mathbf{x}_0$ and including in the local covariance matrix only those points that lie inside a cutoff radius $r_{\rm c}$, which is then varied. 

The main limitation of the global PCA is that it can only detect the correct ID of linearly embedded manifolds (i.e.~linear manifolds $\mathbb{R}^{d} \in \mathbb{R}^{D}$ embedded trivially via rotations and translations), systematically overestimating the ID of curved/non-linearly embedded datasets. 
The mPCA could in principle fix this issue. 
However, PCA only works if the number of samples $N \gtrsim d \log d$, otherwise being inconclusive (see top left panel of Fig. \ref{fig:intro}). 
This is a major drawback in the multiscale case, since to measure the correct ID the cutoff radius $r_{\rm c}$ needs to be small enough, implying that the sampling of the manifold must be dense enough to guarantee that a sufficient number of samples lies inside the subsets $X(\mathbf{x}_0,r_{\rm c})$. 
Another technical issue that makes mPCA difficult to employ for IDE is the
fact that the amplitude of the ``jump'' in the magnitude of the sorted eigenvalues
depends on the data, and the choice of a threshold size is somewhat arbitrary.

The other group of estimators belongs to the so-called \emph{geometric} (or fractal) methods.
Their common ancestor is the correlation dimension (CorrDim) introduced by Grassberger and Procaccia \cite{1983GrassbergerProcacciaMeasuringTheStrangenessOfStrangeAttractors} to measure the fractal dimension of strange attractors in the context of dynamical systems. This estimator is based on the observation that the density of neighbours (also known as correlation integral in the literature) with a given cutoff distance $r$ 
\begin{equation}
\rho_X(r) = \frac{2}{N (N-1)} \sum_{1 \leq \mu < \nu \leq N} \theta \left(r - \lVert\mathbf x^{\mu}-\mathbf x^{\nu}\rVert \right)
\label{FCI}
\end{equation} 
scales as $\rho_X(r) \sim r^d$ for $r \rightarrow 0$ and therefore one can extract the ID by measuring the slope of the linear part of $\rho$ as a function of $r$ in log-log scale, since the relation $d = \lim_{r\rightarrow 0} \log \rho_X (r)/\log r$ holds (see bottom left panel of Fig.~\ref{fig:intro}). 

CorrDim is very effective for the estimation of low IDs ($d \lesssim 10$), whereas it systematically underestimates in the case of manifolds with larger IDs. 
This drawback is well known in the literature \cite{2014CerutiBassisRozzaEtAlDancoAnIntrinsicDimensionalityEstimatorExploitingAngleAndNormConcentration} and is only partially mitigated by more recent and advanced generalizations of CorrDim based on $k$-nearest-neighbors distances \cite{2004LevinaBickelMaximumLikelihoodEstimationOfIntrinsicDimension}. 
The reason why all these algorithms systematically fail for $d \gtrsim 10$ is due to a fundamental limitation of most geometric methods: indeed it is possible to prove \cite{1992EckmannRuelleFundamentalLimitationsForEstimatingDimensionsAndLyapunovExponentsInDynamicalSystems} that the accurate estimation of the ID requires a number of samples $N$ which grows exponentially in the intrinsic dimension $d$. 
As a consequence, one observes a systematic undersampling of the small radius region of the density of neighbors $\rho_X(r)$, as shown for the CorrDim estimator in the bottom left panel of Fig.~\ref{fig:intro}. 

Both mPCA and CorrDim, as well as their more recent generalizations such as DANCo \cite{2014CerutiBassisRozzaEtAlDancoAnIntrinsicDimensionalityEstimatorExploitingAngleAndNormConcentration}, are based on the fundamental fact that, locally, samples in a dataset are effectively drawn uniformly from a $d$-dimensional disk 
$ \{ \pmb{x} \in \mathbb{R}^{d} \, : \, ||x||\leq 1 \} $ linearly embedded in $\mathbb{R}^{D}$.
This is the informal way to state rigorous results based on the tangent space approximation to smooth manifolds and embeddings \cite{2014CerutiBassisRozzaEtAlDancoAnIntrinsicDimensionalityEstimatorExploitingAngleAndNormConcentration,Diaz_2019}.
On one side, local neighbourhoods of large ID datasets ($d \gtrsim 10$) need a number of points $N$ exponential in the ID to be sufficiently sampled; on the other hand, the tangent space approximation needs dense sampling of small patches of the manifold to be used in practice.
This incompatibility of requirements is the so called \emph{curse of dimensionality}, and defines a theoretical limit for all multiscale projective and geometric ID estimators.

In order to try and break the curse of dimensionality,
additional information about the probability distribution of the data must be assumed.
The main ingredient of our spell is the assumption that data are locally isotropic.
This suggests to consider the
average correlation integral for hyperspheres $\{ \pmb{x} \in \mathbb{R}^{d} \, : \, ||x||= 1 \}$.
We will leverage on this observation to develop a novel geometric ID estimator for linearly embedded manifolds,
which departs from the small radius limit of the density of neighbors $\rho_X(r)$ 
and considers this quantity at finite radius $r$.
We show in the following that this method overcomes the extreme undersampling issue caused by the curse of dimensionality and displays a remarkable robustness to non-uniform sampling and noise.
Based on the intriguing features reported above, we propose a multiscale generalization of the estimator, capable of providing the correct ID for datasets extracted from highly curved or multidimensional manifolds.

\begin{figure}
\centering
\includegraphics[width=\columnwidth]{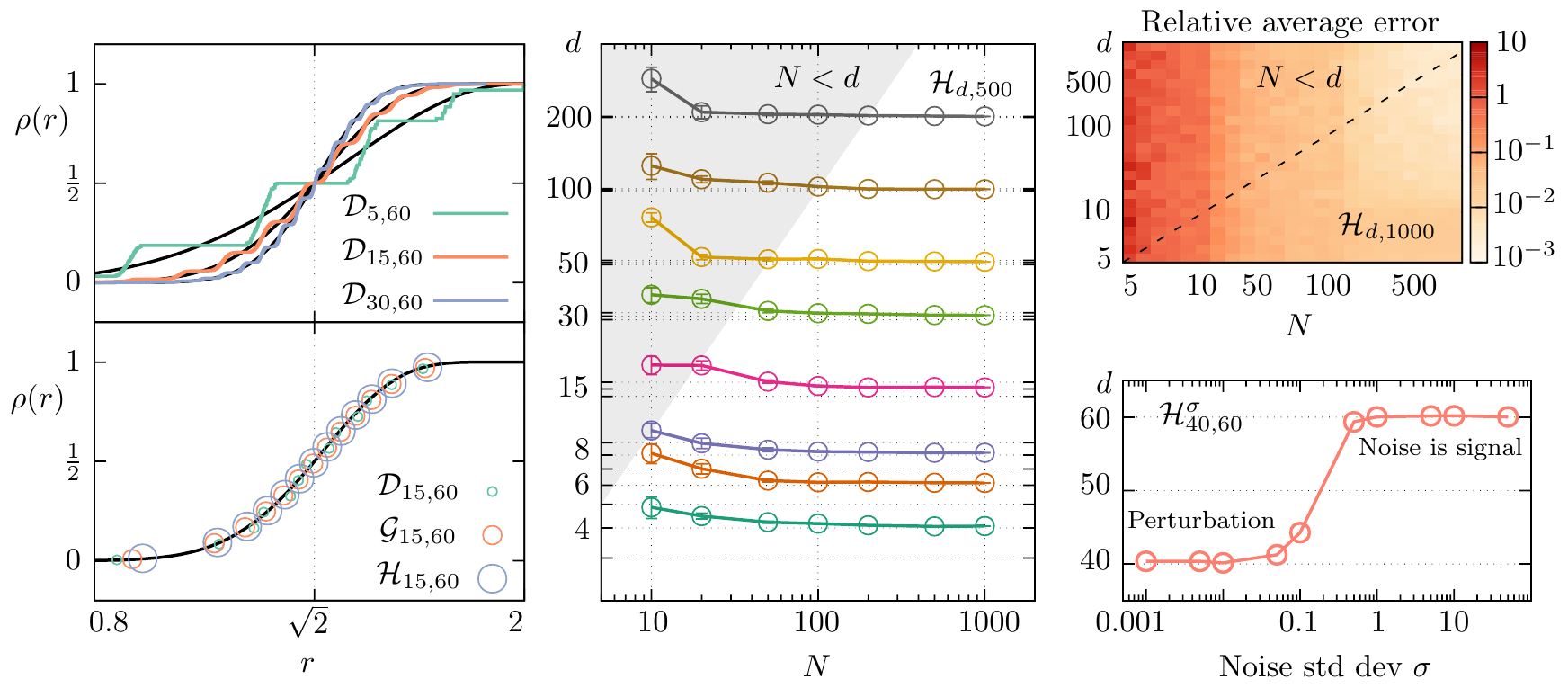}
\caption{\footnotesize
    \textbf{Intrinsic Dimension (ID) estimation is possible in the extreme undersampled regime for arbitrarly large ID with the Full Correlation Integral (FCI) estimator, in the case of linearly embedded and slightly curved manifolds, possibily non uniformly sampled and with noise.}
    \textbf{(Top left)} We show the density of neighbours $\rho$ of preprocessed (centered and normalized) data (number of samples $N=500$) extracted from $\{0,1\}^{d}$ linearly embedded in $D=60$ dimensions ($\mathcal{D}_{d,60}$), for $d=(5,15,30)$. 
    We are able to efficiently extract the correct ID even though this is a highly non-uniformly sampled dataset, whose $\rho$ displays manifold-dependent features (in this case, step-like patterns).
    Moreover, we observe that, as we increase $d$, the density of neighbours of this dataset quickly converges to our functional form.
    It is worth noticing that the whole functional form \eqref{eq:FCI} is needed for the fit; in fact, a local fit of the slope of $\rho$ at half-height would result in an incorrect ID estimation.
    \textbf{(Bottom left)} We show the density of neighbours $\rho$ of preprocessed data ($N=500$) extracted uniformly from $\{0,1\}^{d}$, $[0,1]^{d}$ and from $\mathbb{R}^{d}$ with multivariate gaussian distribution, for $d=15$ and linearly embedded in $D=60$ dimensions.
    All plot lines are compatible with the same functional form \eqref{eq:FCI}, pointing to an intriguing manifestation of "universality" for high-dimensional data.
    \textbf{(Center)} To highlight the predictive power of the FCI method for a broad spectrum of dimensionalities (ranging from $d=4$ to $d=200$), we exhibit the estimated ID versus the number of sample points $N$ for the linearly embedded hypercube $\mathcal{H}_{d,500}$.
    Error bars are computed by averaging over 10 samples for each pair $(N,d)$.
    \textbf{(Top right)} We asses quantitatively the predictive power of the FCI method by computing the average relative error $ |(d_{\rm est}-d)/d| $ (over 20 random instances) of the estimated ID in the range $5 \leq d \leq 1000$, $5\leq N \leq 1000$.
    We observe that at $N\sim100$ we have an error of the order of $1\%$ almost independently on the ID, and that ID estimation is possible also in the extreme undersampled $N<d$ regime.
    \textbf{(Bottom right)} The FCI method estimates the correct ID even when the data are corrupted by noise. 
    Here we consider a linearly embedded hypercube dataset $\mathcal{H}_{40,60}$ and add on the top of that a $60$-dimensional gaussian noise of standard deviation $\sigma$.
    We observe a sharp transition in the estimated ID between the regime in which the noise is a perturbation ($\sigma \lesssim 0.1$ and $d_{\rm est}=40$) and the regime in which the noise covers the signal ($\sigma \gtrsim 0.2$ and $d_{\rm est}=60$).}
\label{fig:global}
\end{figure}


\subsection*{Full Correlation Integral (FCI) Estimator} 

The tangent space approximation suggests that a special role in the IDE problem is played by uniformly sampled $d$-dimensional disks linearly embedded. The \emph{average} correlation integral for the boundary of this manifold, which is the $(d-1)$-dimensional sphere of radius $r_{\rm s}$, can be analytically evaluated as (see Materials and methods) 
\begin{equation}
\label{eq:FCI}
\overline{\rho_S(\bar{r})} = \frac{1}{2} + \frac{\Omega_{d-1}}{2 \Omega_{d}}  (\bar{r}^{2}-2) _{2}F_{1}\left(\begin{array}{c} \frac{1}{2} , 1-\frac{d}{2} \\ \frac{3}{2} \end{array} \bigg\rvert \, (\bar{r}^{2}-2)^{2} \right) \,,
\end{equation}
where $_{2}F_{1}$ is the $(2,1)$-hypergeometric function, $\Omega_d$ is the $d$-dimensional solid angle and $\bar{r} = r/r_{\rm s}$ is the adimensional cutoff radius. 
We take \eqref{eq:FCI} as the definition of the \emph{full} correlation integral, to stress that we work away from the small radius limit employed for CorrDim. 

It is worth noticing that the FCI has a sigmoidal shape which is steeper as the ID grows (see for instance the black lines in the top left panel of Fig. \ref{fig:global}). 
This observation translates into a simple exact algorithm to determine the ID $d$ of linearly embedded spherical datasets, by performing a non-linear regression of the empirical density of neighbours using the FCI in \eqref{eq:FCI}. More in general, this protocol is exact for linearly embedded Euclidean spaces sampled with a rotational invariant probability distribution, by projecting onto the unit sphere and adding one to the ID estimated on this new dataset.    
We summarize our FCI estimator in two steps:
\begin{enumerate}
\item compute the center of mass $\mathbf b$ of the empirical data as $\mathbf b = 1/N \sum_{\mu} \mathbf x^{\mu}$ and translate each datapoint by this quantity, so that the resulting dataset is centered at the origin. 
Then normalize each sample;
\item measure the empirical correlation integral of the dataset as a function of the radius $r$ and perform a non-linear regression of this empirical density of neighbours using the FCI in \eqref{eq:FCI} as the non-linear model and $d$ and $r_{\rm s}$ as the free parameters;
as the normalization step artificially removes one degree of freedom, 
increase the estimated ID by one.
\end{enumerate}
We discuss the technical details regarding the fitting protocol in the Methods.

\subsection*{Robustness of the FCI estimator}

We now provide strong numerical evidence that the FCI estimator goes well beyond the exact results summarized above by testing it on multiple synthetic non-spherical datasets.

First, we notice that manifold-dependent features tend to disappear from the empirical correlation integral as the ID grows, quickly converging to the FCI prediction for angularly uniform data. In the top left panel of Fig.~\ref{fig:global}, we highlight this effect by showing the empirical correlation integral for three single instances ($d=5,15,30$) of the dataset uniformly drawn from $\{0, 1\}^d$, linearly embedded in $D = 60$ dimensions ($\mathcal D_{d,60}$). 
Among the many datasets breaking rotational invariance that we have investigated, this is one of those where manifold-dependent features are more pronounced: we observe a ladder-type pattern that quickly disappears as the ID grows from
$5$ to $30$.
Nonetheless, even in the low-dimensional case, where the steps may affect the ID estimation, the FCI method works.
We stress that in many relevant cases, including linearly embedded and uniformly sampled hypercubes ($\mathcal H_{d,D}$), deviations from \eqref{eq:FCI} are negligible even in low dimension. In the bottom left panel of Fig.~\ref{fig:global}, we 
substantiate this point
by comparing the empirical FCI the three datasets $\mathcal D_{15,60}$, $\mathcal H_{15,60}$, and a rotationally invariant dataset sampled with radial Gaussian distribution $\mathcal G_{15,60}$. 

The FCI estimator shares some similarities with the one recently introduced in \cite{2016GranataCarnevaleAccurateEstimationOfTheIntrinsicDimensionUsingGraphDistancesUnravelingTheGeometricComplexityOfDatasets} by Granata and Carnevale; here the authors use a derivative of our empirical correlation integral as the non-linear model to fit the mid-height section of the curve.
Compared to the method of Granata and Carnevale, the FCI estimator has two additional major strengths. 
First, the normalization procedure sets a common typical scale for all datasets,
making the comparison with \eqref{eq:FCI} straightforward.
Second, and perhaps more importantly, 
our non-linear fit is performed by taking into account the whole functional form in \eqref{eq:FCI}, and not only the mid-height local portion of the empirical FCI; as the top left panel of Fig.~\ref{fig:global} shows, manifold-dependent features are hardly avoided if the fit is performed locally.

Our estimator is robust to extreme undersampling and to noise, as highlighted in the center and right panels of Fig.~\ref{fig:global} for the $\mathcal H_{d,D}$ dataset. In a broad ID range (up to $d=10^3$), we provide accurate estimates even in the regime $N < d$ with a relative average error that decays quickly with the number of samples $N$ (almost independently  from the ID); as a matter of fact, for $N=100$ the error is already below one percent. 
The method is also particularly robust to Gaussian noise, showing a sharp crossover between the regime where the ID is correctly retrieved to a phase where the noise covers the signal. 

So far, we have applied and verified the performance of our algorithm only on linearly embedded manifolds. 
More in general, 
we have verified that
the FCI estimator correctly identifies the ID even in the case of simple non-linear polynomial embeddings or of slightly curved manifolds. 
However, the method presented above is \emph{global}, thus it
is expected to fail on manifolds with high intrinsic curvature. As in the case of the global PCA \cite{Little_2009, 2017LittleMaggioniRosascoMultiscaleGeometricMethodsForDataSetsIMultiscaleSVDNoiseAndCurvature}, 
we can overcome this issue by providing a suitable multiscale generalization of the FCI estimator.

\begin{figure}
\centering
\includegraphics[width=\columnwidth]{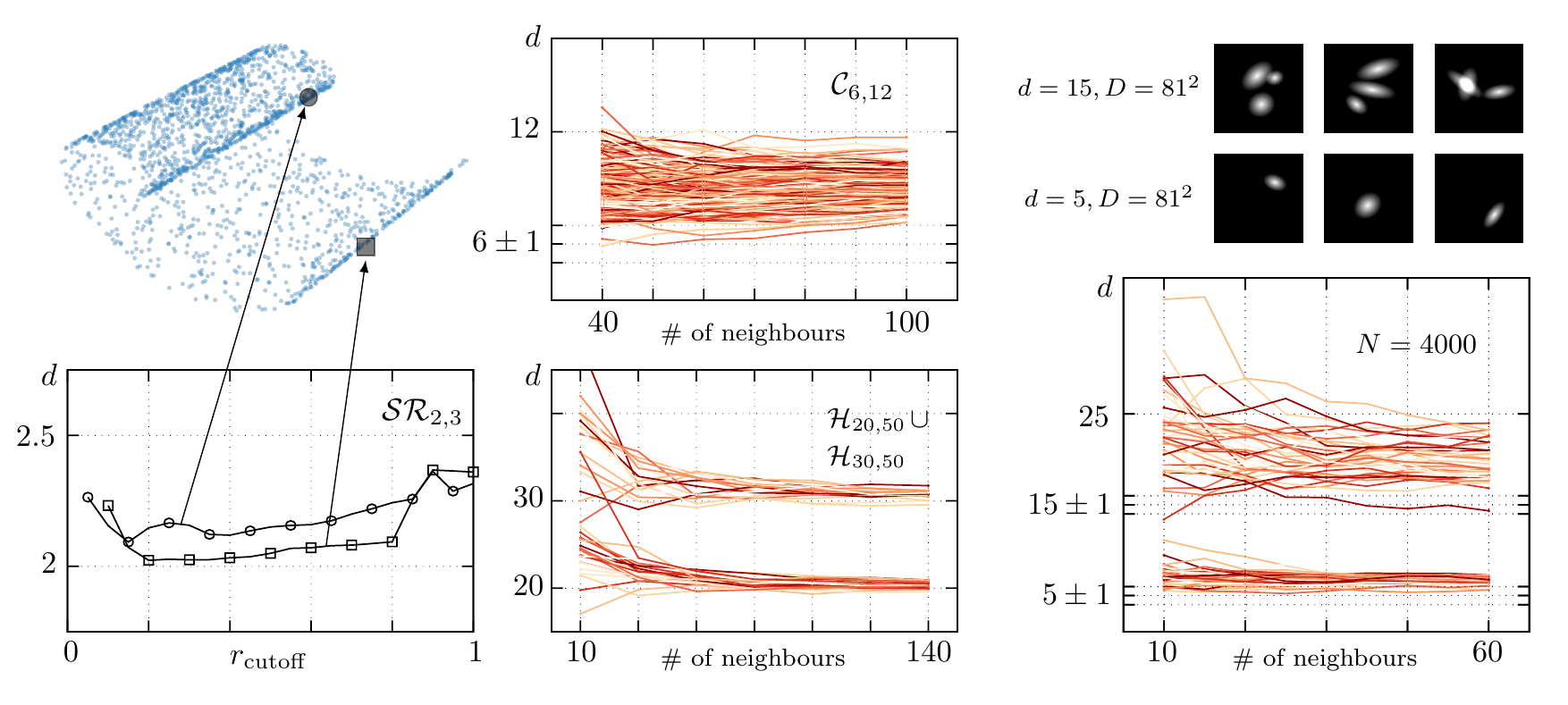}
\caption{\footnotesize
    \textbf{The multiscale generalization of the FCI estimator provides a state-of-the-art tool to tackle the ID estimation for complex datasets with multidimensional features and high intrinsic curvature.}
    The multiscale FCI method selects single points in the dataset and their neighbours at a fixed maximum distance $r_{\rm cutoff}$, which is then varied.
    The FCI estimator is then used on each neighbourhood, giving an estimation for a local ID $d_{\rm est}(r_{\rm cutoff})$. 
    Crucially, the robustness to extreme local undersampling of the FCI estimator allows to shrink the radius of the neighbourhoods $r_{\rm cutoff}$, giving a reliable estimate of local IDs, that appear as pronounced plateaux in the $d_{\rm est}$ vs $r_{\rm cutoff}$ plot.
    Alternatively, one can look at $d_{\rm est}(n)$, where $n$ is the number of nearest neighbours used in the estimation, keeping more control on the number of points used in the local estimation.
    \textbf{(Left)} We present an illustrative application of the multiscale FCI method in the case of the Swiss Roll dataset ($N=2000$) for two particular samples.
    Both samples hint to the correct ID estimation $d_{\rm est}=2$.
    We observe that the sample extracted from the highly curved inner region of the Swiss Roll provides an overestimated ID whereas the sample from the outer and flatter region exhibits a plateau at the correct ID.
    This suggests to use the minimum $d_{\rm est}$ reached as an estimator for the true ID, but a more careful analysis is needed. 
    \textbf{(Top center)} We apply the multiscale FCI method to a highly curved manifold $\mathcal{C}_{6,12}$ ($N=10000$) introduced in \cite{2005HeinAudibertIntrinsicDimensionalityEstimationOfSubmanifoldsInRd}, challenging to all ID estimators (see also Materials and methods for its parametric defintion).
    The local $d_{\rm est}$ spans the range between the true ID $d_{\rm est} \sim 6$ to the embedding dimension $D\sim12$.
    \textbf{(Bottom center)} We show the multiscale FCI analysis on a multidimensional manifold which is built as the union of two intersecting hypercubes datasets $\mathcal{H}_{20,50}$ and $\mathcal{H}_{30,50}$, each one consisting of $N=1000$ samples. 
    At small number of neighbours, we observe spurious effects due to vary sparse sampling of the neighbourhoods; quick convergence to the true IDs is then observed.
    \textbf{(Right)} As a last validation test, we generate an artifical dataset of bitmap images with multiple "blobs" with five degrees of freedom each (see Materials and methods), that we use as a proxy for curved manifolds of transformations of high contrast images. The multiscale analysis works nicely in either the one ($d=5$) and three ($d=15$) blob cases, although for multiple blobs we observe that the high curvature reflects in band that spans more than ten dimensions.      
}
\label{fig:local}
\end{figure}

\subsection*{Multiscale FCI estimator}

We perform a multiscale analysis
of the FCI estimator by selecting a random sample $\mathbf x_0$ in the dataset and a cutoff radius $r_{\rm c}$. 
We then apply the FCI estimation method to the set of points whose distance from $\mathbf x_{0}$ is less then $r_{\rm c}$.
In this way, we obtain a local ID estimate $d_{\mathbf x_0} (r_{\rm c})$ that depends on the cutoff radius. Varying $r_{\rm c}$ and $\mathbf x_0$, we obtain a family of curves that describes the local ID of the dataset at different scales (see also left panel in Fig.~\ref{fig:local}). Another possibility to perform the multiscale analysis is to control the number of neighbors used for the local ID estimation; 
this parameter is clearly in one-to-one correspondence with the cutoff radius.

First, we look at a paradigmatic curved manifold studied in the literature, the Swiss roll $\mathcal {SR}_{2,3}$ (left panel in Fig.~\ref{fig:local}). 
In general, we observe three regimes: for very small cutoff radius, the local estimation by the FCI is not reliable due to the extreme scarcity of neighbouring points ($N \lesssim 20$). 
For very large cutoff radius, the local ID converges to the global FCI estimation, as more and more samples lie inside the cutoff radius. 
In between, if the manifold is sampled densely enough, we observe a plateau.
In the central bottom panel of Fig. \ref{fig:local}, we show two representative samples of the $\mathcal {SR}_{2,3}$ dataset ($N = 2000$), chosen from regions with very different curvatures. 
When the curvature is large, the height of the plateau identifies an overestimated ID, 
which drifts towards the embedding dimension as expected.
On the other hand, when the curvature is small, the correct ID is identified.
This observation suggests 
that the best estimator for the multiscale analysis could
be the minimum ID identified in the plateau region.
We use this heuristic in the tests below, but we leave a more
detailed investigation of the multiscale method for future work.

Now we move to multidimensional datasets and non-trivially curved manifolds, which are considered challenges for state-of-the-art ID estimators.
In the bottom center panel of Fig.~\ref{fig:local}, we show the multiscale analysis for an instance of the dataset $\mathcal H_{20,50} \cup H_{30,50}$ ($N=1000+1000$), representing two intersecting hypercubes with different IDs. Two plateaux at $d=20$ and $d=30$ are clearly visible and allow to infer the multidimensionality of the dataset (in this case the measured local dimension is displayed as a function of the number of neighbors used for the estimation). 
In the top center panel of Fig.~\ref{fig:local}, we display the same analysis for an instance of the dataset $\mathcal C_{6,12}$ ($N=2500$), first introduced in \cite{2005HeinAudibertIntrinsicDimensionalityEstimationOfSubmanifoldsInRd} and considered a challenging dataset for ID estimation for its high intrinsic curvature. 
Although a thorough comparison of the performance of our algorithm with state-of-art ID estimators is out of the scope of our investigation (see \cite{2014CerutiBassisRozzaEtAlDancoAnIntrinsicDimensionalityEstimatorExploitingAngleAndNormConcentration} or \cite{2015AmsalegChellyFuronEtAlEstimatingLocalIntrinsicDimensionality} for nice recent meta-analysis), we observe that our prediction $d\simeq 5.9$ is pretty accurate (state-of-the-art estimators such as DANCo find $d\simeq 6.9$ on the $\mathcal C_{6,12}$ dataset sampled in the same conditions).   

This excellent accuracy on highly curved manifolds, combined --\emph{at the same time}-- with the removal of the well known underestimation issue common to all geometric methods, provides two major advantages of our algorithm over other traditional schemes used for ID estimation and suggests to test and validate it on the manifolds of global transfomations (e.g. translations, rotations, dilations) generated by high contrast images, relevant in the context of invariant object recognition. 
These manifolds often feature high local curvature, since even infinitesimal transformations produce almost orthogonal tangent spaces \cite{Bengio:2004:NMT:2976040.2976057,Bengio2}.

As a more demanding test for our multiscale FCI estimator, we report a preliminary investigation of a manifold (artificially generated) that belongs to this class, where we can keep under control the intrinsic dimensionality (see right panels of fig. \ref{fig:local}). 
We consider bitmap images with multiple blobs (possibly overlapping) with five degrees of freedom each (two for translations, one for rotation, two for asphericity and dilation, see also Materials and methods). 
Even here the multiscale analysis provides a reliable indicator, as we can convince by looking at the one and three blob cases ($d=5$ and $d=15$ respectively). 
It is worth noticing that whereas single points are affected by the high curvature of the manifold (resulting in higher ID estimations), the minimum of the plateaux works again nicely as the estimator of the correct ID. 

\section*{Comparison with existing methods}

As a preliminary quantitative comparison with the existing ID estimators, we analysed all datasets presented in Figure~\ref{fig:local} with a selection of geometric and projective estimators available in the literature.

\begin{figure}
    \begin{subfigure}{0.62\textwidth}
        \small
        \begin{tabular}{r|c|c|c|c|c}
            \hline
            Estimator & $\mathcal{SR}_{2,3}$ & \makecell{$\mathcal{H}_{20,50}$ \\ $\cup$ \\ $\mathcal{H}_{30,50}$} & $\mathcal{C}_{6,12}$ & $\mathcal{B}_{5,81^{2}}$ & $\mathcal{B}_{15,81^{2}}$ \\
            \hline
            \hline
            CorrDim \cite{1983GrassbergerProcacciaCharacterizationOfStrangeAttractors} 
                      & \color{OliveGreen}{1.98} & \color{Maroon}{12.53} & \color{OliveGreen}{5.93} & \color{OliveGreen}{5} & \color{Maroon}{13.5} \\
            Takens \cite{Takens_1985} 
                      & \color{OliveGreen}{1.97} & \color{Maroon}{12.01} & \color{OliveGreen}{5.77} & \color{Maroon}{N.A.} & \color{Maroon}{N.A.} \\
            Hein et al. \cite{2005HeinAudibertIntrinsicDimensionalityEstimationOfSubmanifoldsInRd} 
                      & \color{OliveGreen}{2} & \color{Maroon}{13} & \color{OliveGreen}{6} & \color{Maroon}{N.A.} & \color{Maroon}{N.A.} \\
            \hline
            \hline
            PCA
                      & \color{Maroon}{3} & \color{OliveGreen}{20 \& 30} & \color{Maroon}{12} & \color{Maroon}{40} & \color{Maroon}{40} \\
            mPCA \cite{2017LittleMaggioniRosascoMultiscaleGeometricMethodsForDataSetsIMultiscaleSVDNoiseAndCurvature}
                      & \color{Maroon}{3} & \color{OliveGreen}{20 \& 30} & \color{Maroon}{[9,12]} & \color{Maroon}{[2,10]} & \color{Maroon}{[6,31]}\\
            \hline
            \hline
            Multiscale FCI & \color{OliveGreen}{2} & \color{OliveGreen}{20 \& 30} & \color{OliveGreen}{6} & \color{OliveGreen}{5} & \color{OliveGreen}{15} \\
            \hline
        \end{tabular}
    \end{subfigure}
    \begin{subfigure}{0.3\textwidth}
        \includegraphics[width=6cm]{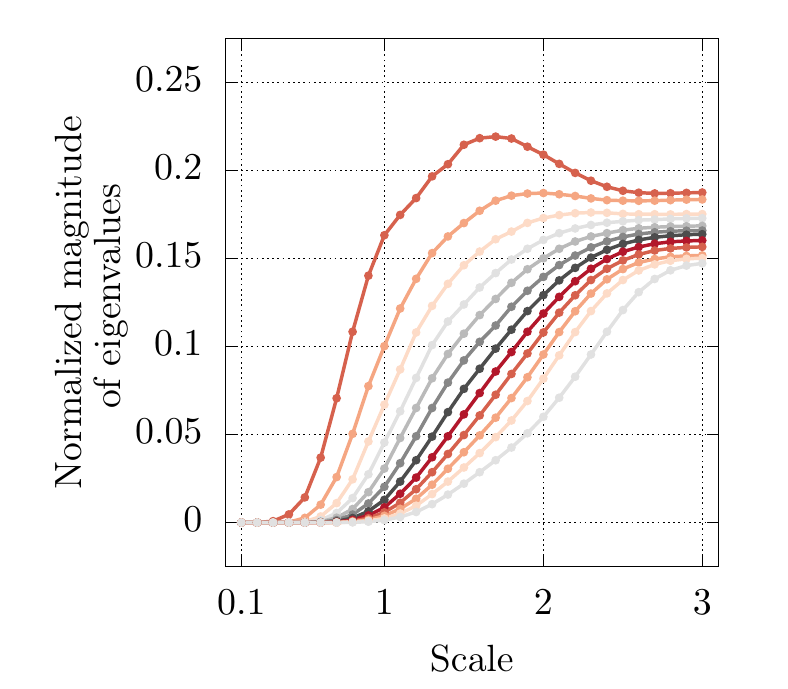}
    \end{subfigure}
    \caption{\footnotesize
        \textbf{Comparison between ID estimators on curved and multidimensional datasets.}
        Geometrical methods fail on high-ID datasets, even if the embedding is linear.
        Global PCA behaves complementarily, retrieving correctly the ID in this case, but losing predictivity on curved datasets.
        This issue is not fixed by perfoming multiscale PCA, since we often lack a clear signature for estimating the ID, e.g.~a gap in the magnitude of the sorted eigenvalues.
        Even if we use the less stringent criterion (often used in the literature, see for instance \cite{2017LittleMaggioniRosascoMultiscaleGeometricMethodsForDataSetsIMultiscaleSVDNoiseAndCurvature}) of identifying the ID as the minimum number of eigenvalues such that their mass $\sum_{i=1}^{\textrm{ID}}\lambda_i/\sum_{i=1}^{\textrm{D}}\lambda_i$ is larger than $0.95$, we lack a signature of \emph{persistence} as in the case of our estimator (the plateau as a function of the cutoff radius).
        In the right panel we plot the twelve averaged eigenvalues of the correlation matrix of the $\mathcal{C}_{6,12}$ manifold as a function of the cutoff scale (the average is performed over all the different balls of the same radius centered around each point) to highlight this issue.
        No evidence of the correct ID can be found using the common criteria reported above.
        Where \emph{not available} (N.A.) is reported, the code of \cite{2005HeinAudibertIntrinsicDimensionalityEstimationOfSubmanifoldsInRd} returned either 0 or infinity.
        We expect however that the results would be very similar to those obtained with CorrDim.}
        \label{fig:comparison}
\end{figure}

On the side of geometric estimators, we studied the Takens estimator \cite{Takens_1985}, the correlation dimension \cite{1983GrassbergerProcacciaCharacterizationOfStrangeAttractors} and the estimator introduced in \cite{2005HeinAudibertIntrinsicDimensionalityEstimationOfSubmanifoldsInRd}, using the code available at \url{https://www.ml.uni-saarland.de/code/IntDim/IntDim.htm}.
The results of the analysis are presented in Figure~\ref{fig:comparison}: geometric methods are reliable in low dimension, independently from the curvature of the manifold under exam, whereas for $\textrm{ID} > 10$ they start to experience the familiar underestimation issue.

Principal component analysis, on the contrary, identifies correctly the ID even in high dimension, as long as the manifold is linearly embedded. In particular, in the case of the multidimensional dataset examined here, we are able to identify two different gaps in the magnitude of the sorted eigenvalues at the correct IDs. On curved manifolds instead, global PCA overestimates dramatically as expected. 
This problem should be in principle fixed by performing a multiscale analsyis (mPCA) as proposed in \cite{2017LittleMaggioniRosascoMultiscaleGeometricMethodsForDataSetsIMultiscaleSVDNoiseAndCurvature}. However, for the curved manifolds considered here, this does not lead to a significant improvement in the ID estimation. The irrelevant ($\textrm{D} - \textrm{ID}$) eigenvalues of the correlation matrix should go much faster to zero than the remaining relevant ones when the cutoff radius is reduced. In practice, we are not able to identify a clear signature of this phenomenon and we can only establish loose bounds on the ID (see also the Figure~\ref{fig:comparison}). 

\section*{Discussion}

In this manuscript we introduced the FCI estimator for the ID of spherically sampled, linearly embedded datasets, showing that it is robust to noise and non-idealities and more importantly that it works effectively in the extreme undersampled regime ($N<d$).

We performed a multiscale analysis of the FCI estimator on challenging datasets, featuring high curvature and multidimensionality, showing that we can extract the correct ID as the minimum local $d_{\rm est}$.
Further work will be needed to fully explore this observation and to construct a proper multiscale estimator.

    We performed a preliminary comparative analysis of our estimation framework against some representative geometric and projective ID estimators, and found that the multiscale FCI can provide reliable predictions in a variety of different regimes, while the other estimators tipically excel only under specific conditions (small dimension for geometric estimators, small curvature for projective estimators). 

We leave open to future investigations the analysis of high dimensional manifolds of high contrast images taken from the Machine Learning literature, as well as the possibility of combining our estimator with state-of-the-art techniques for dimensional reduction and manifold learning, or, even more ambitiously, to elaborate on it in order to propose a novel more effective toolbox for these tasks.

Beyond these applications, ID estimation has been very recently used \cite{2019arxiv190512784A, 2019arxiv190600443R} by the Machine
Learning community as a tool to understand how deep neural networks transform and compress information in their hidden layers. 
Here the authors observe that the range of IDs of many training sets (such as Fashion-MNIST and CIFAR-10) processed through the hidden layers of a DNN is between 10 and 100.
This is the typical regime where our ID estimator overcomes standard methods, so that it would be interesting to use it to reproduce these analysis.

\section*{Materials and methods}

\subsection*{Average correlation integral for uniformly sampled hyperspheres}

Here we derive the average correlation integral Eq.~\eqref{eq:FCI} for a dataset uniformly sampled from the hypersphere $\mathcal S^{d}$ or radius $r_{s}$, i.e. 
\begin{equation} \label{eq:FCIfull}
    \begin{split}
        \overline{\rho_S(r)} 
        &= \left\langle \frac{2}{N(N-1)} \sum_{1\leq i < j \leq N} \theta(r - || \pmb{x}_{i} - \pmb{x}_{j} ||) \right\rangle\\
        &= \frac{2}{N(N-1)} \sum_{1\leq i < j \leq N} \left\langle \theta(r^{2} - || \pmb{x}_{i} - \pmb{x}_{j} ||^{2}) \right\rangle\\
        &= \frac{2}{N(N-1)} \sum_{1\leq i < j \leq N} \int_{\mathcal{S}^{d}} d\pmb{x}_{i} d\pmb{x}_{j} \mu_{\rm uniform}(\pmb{x}_{i}) \mu_{\rm uniform}(\pmb{x}_{j}) \theta(r^{2} - || \pmb{x}_{i} - \pmb{x}_{j} ||^{2}) \\
&=\int_{\mathcal S^{d}} d\pmb{\alpha}\, \frac{\Omega_{d}(\pmb{\alpha})}{\Omega_{d}} \, d\pmb{\beta}\, \frac{\Omega_{d}(\pmb{\beta})}{\Omega_{d}} \theta \left(  r^{2} - || \pmb{x}(\pmb{\alpha}) - \pmb{x}(\pmb{\beta}) ||^{2}  \right) \, ,
    \end{split}
\end{equation}
where $\pmb \alpha$ and $\pmb \beta$ are spherical coordinates (i.e. $\alpha_{i} \in (0,\pi)\,\, \forall i=2\dots d$ and $\alpha_{1}\in(0,2\pi)$ and the same for $\beta_{i}$),     $\Omega(\pmb{\alpha})=\sin(\alpha_{2}) \sin^{2}(\alpha_{3}) \dots \sin^{d-1}(\alpha_{d})$ is the $d$-dimensional spherical volume element and $\mathbf{x}(\cdot)$ is the function that converts spherical coordinates into $(d+1)$-dimensional Euclidean coordinates on the sphere of radius $r_{s}$.
$\Omega_{d} = \int_{S_{d}} d\pmb{\alpha} \Omega_{d}(\pmb{\alpha})$ is the $d$-dimensional solid angle.
The integral can be evaluated by choosing the spherical coordinates $\pmb \beta$ such that their azimuth axis is in the direction of $\pmb \alpha$, so that
\begin{equation}
    ||x(\pmb{\alpha}) - x(\pmb{\beta})||^{2} = 2 r^{2}_{s} (1-\cos(\beta_{d})) \, .
\end{equation}
The integrals in $\pmb \alpha$ and in $\beta_{1}\dots\beta_{d-1}$ are trivial, giving 
\begin{equation}
    \begin{split}
        \overline{\rho_S(\bar{r})} &= \frac{\Omega_{d-1}}{\Omega_{d}} \int_{0}^{\arccos\left( 1-\frac{\bar{r}^{2}}{2} \right)} d\beta_{d} \sin^{d-1}(\beta_{d}) \\
                                   &= \frac{1}{2} + \frac{\Omega_{d-1}}{2 \Omega_{d}}  (\bar{r}^{2}-2) _{2}F_{1}\left(\begin{array}{c} \frac{1}{2} , 1-\frac{d}{2} \\ \frac{3}{2} \end{array} \bigg\rvert \, (\bar{r}^{2}-2)^{2} \right) \,,
         \, 
    \end{split}
\end{equation}
where $\overline{r}=r/r_{s}$.

\subsection*{Empirical full correlation integral and fitting procedure}

The empirical full correlation integral is easily computed in \texttt{Mathematica 12} using the following one-liner:
\begin{alltt}
    Module[\{dists = Sort@(Norm[#[[1]] - #[[2]]] & /@ Subsets[sample, \{2\}])\}, \\
    Transpose[\{dists, N@(Range[Length[dists]] - 1)/Length[dists]\}]]
\end{alltt}

The non-linear fit for the FCI estimator was performed using the default function \texttt{FindFit}, with $d$ and $r_{s}$ as free parameters.
The empirical full correlation integral was preprocessed before the application of \texttt{FindFit} by extracting a \texttt{RandomSample} of $\min\left(1000,\frac{N(N-1)}{2}\right)$ of its points to speed-up the fitting procedure.

\subsection*{Description of the datasets}

In this section we briefly describe the datasets used in the presented numerical simulations.
In the following, a linear embedding is the map $\iota: \mathbb{R}^{d}\rightarrow\mathbb{R}^{D}$ that appends $D-d$ zeros to its argument, and then rotates it in $\mathbb{R}^{D}$ by a randomly chosen rotation matrix.

\begin{description}
    \item[$\mathcal{D}_{d,D}$: ] uniform sampling of $\{0,1\}^{d}$, linearly embedded.
    \item[$\mathcal{G}_{d,D}$: ] sampling of $\mathbb{R}^{d}$ with the multivariate Gaussian distribution of covariance matrix $\mathds{1}$ and null mean, linearly embedded.
    \item[$\mathcal{H}_{d,D}$: ] uniform sampling of $[0,1]^{d}$, linearly embedded.
    \item[$\mathcal{C}_{d,2d}$:] uniform sampling of $[0,2\pi]^{d}$, embedded with the map
        \begin{equation*}
            \phi(x_{1}\dots x_{d})= (x_{2}\cos(x_{1}), x_{2}\sin(x_{1})\dots x_{1}\cos(x_{d}), x_{1}\sin(x_{d}))\, .
        \end{equation*}
    \item[$\mathcal{SR}_{2,3}$: ] uniform sampling of $[0,1]^{d}$, embedded with the map 
        \begin{equation*}
            \phi(x,y) = (x \cos(2\pi y), y, x \sin(2\pi y)) \, .
        \end{equation*}
    \item[$\mathcal{B}_{5n,81^{2}}$: ] dataset of high-contrast bitmap images (81$\times$81 pixel) of $n$ blobs. $n=1$ images generation is described in the following section. $n>1$ images are generated by summing $n$ different $n=1$ images.
\end{description}

\subsection*{High-contrast images datasets}

The high-contrast images shown in Figure~\ref{fig:local} (right panel) were generated by assigning to each pixel of a $l\times l$ bitmap the following value $v_{i,j}$:
\begin{equation}
    \begin{split}
        a_{i,j} &=  \cos{\theta} (j-dx) + \sin{\theta} (i+dy)\\
        b_{i,j} &= -\sin{\theta} (j-dx) + \cos{\theta} (i+dy)\\
        v_{i,j} &= 1 - \sqrt{\frac{a_{i,j}^{2}+e^{2}b_{i,j}^{2}}{(1+e^{2})s^{2}}} \, 
    \end{split}
\end{equation}
with parameters
\begin{center}    
\begin{tabular}{|r|l|l|}
    \hline
    $l$         & side of the bitmap       & fixed at 81                        \\ 
    $\Delta x$  & horizontal translation   & uniform in $(-20,20)$\\
    $\Delta y$  & vertical translation     & uniform in $(-20,20)$\\
    $s$         & size                     & uniform in $(1,3)$\\
    $e$         & eccentricity             & uniform in $(5,10)$\\
    $\theta$    & angle of the major axis  & uniform in $(-\pi/2,\pi/2)$        \\
    \hline
\end{tabular}
\end{center}

Morover, any pixel of value less then 0.01 was manually set to 0 to increase the contrast of the image.

\bibliographystyle{unsrt}


\section*{Acknowledgments}
P.R.  acknowledges  funding  by the European Union through the H2020 - MCIF Grant No. 766442.

\end{document}